\def\year{2021}\relax
\documentclass[letterpaper]{article} % DO NOT CHANGE THIS

\usepackage{aaai21}  % DO NOT CHANGE THIS
\usepackage{times}  % DO NOT CHANGE THIS
\usepackage{helvet} % DO NOT CHANGE THIS
\usepackage{courier}  % DO NOT CHANGE THIS
\usepackage[hyphens]{url}  % DO NOT CHANGE THIS
\usepackage{graphicx} % DO NOT CHANGE THIS
\usepackage{natbib}  % DO NOT CHANGE THIS AND DO NOT ADD ANY OPTIONS TO IT
\usepackage{caption} % DO NOT CHANGE THIS AND DO NOT ADD ANY OPTIONS TO IT
\usepackage{amsfonts}       % blackboard math symbols
\usepackage{nicefrac}       % compact symbols for 1/2, etc.
\usepackage{microtype}      % microtypography
\usepackage{amsmath}
\usepackage{amssymb}
\usepackage{subcaption}
\usepackage{algorithm}
\usepackage{algorithmic}
\usepackage{textcomp}
\usepackage{booktabs}

\title{TextGAIL: Generative Adversarial Imitation Learning for Text Generation}

\author{
    Qingyang Wu \textsuperscript{\rm 1},
    Lei Li \textsuperscript{\rm 2},
    Zhou Yu \textsuperscript{\rm 1}, \\
}

\affiliations{
    \textsuperscript{\rm 1}University of California, Davis, \textsuperscript{\rm 2}ByteDance, \\
    \{wilwu, joyu\}@ucdavis.edu, lileilab@bytedance.com \\
}

\begin{document}

\maketitle

\begin{abstract}

    Generative Adversarial Networks (GANs) for text generation have recently received many criticisms, as they perform worse than their MLE counterparts \cite{Caccia2020Language,DBLP:conf/naacl/TevetHSB19,DBLP:journals/corr/abs-1806-04936}.
    We suspect previous text GANs' inferior performance is due to the lack of a reliable guiding signal in their discriminators.
    % Ours
    To address this problem, we propose a generative adversarial imitation learning framework for text generation that uses large pre-trained language models to provide more reliable reward guidance.
    % Details
    As previous text GANs suffer from high variance of gradients,
    we apply contrastive discriminator, and proximal policy optimization (PPO) to stabilize and improve text generation performance.
    % Results
    For evaluation, we conduct experiments on a diverse set of unconditional and conditional text generation tasks.
    Experimental results show that TextGAIL achieves better performance in terms of both quality and diversity than the MLE baseline.
    We also validate our intuition that TextGAIL's discriminator demonstrates the capability of providing reasonable rewards with an additional task.\footnote{Code is available at https://github.com/qywu/TextGAIL}

\end{abstract}

\section{Introduction}

% The background and Problem
% Text generation is gaining popularity in research recently. 
Automatic text generation has been used in tremendous applications such as machine translation, question answering, and dialog system.
The most widely used approach for neural text generation is to maximize the probability of the target text sequence \cite{DBLP:conf/nips/BengioDV00}, which is also referred to as maximum likelihood estimation (MLE).
%% Exposure bias
However, MLE suffers from the exposure bias problem which is due to the discrepancy between training and inference.
%% What is exposure bias
During training, the model is trained on the ground truth, but during inference, the model needs to autoregressively predict the next word conditioned on its own previously generated words.
%% Why it is bad
This discrepancy hurts generalization of unseen data and leads to lower quality of generated text \cite{DBLP:journals/corr/abs-1908-04319, DBLP:conf/emnlp/WisemanR16}. 
Therefore, solving the exposure bias problem becomes an promising approach to improve text generation quality.
% which becomes an central problem in neural text generation.
% when a maximum likelihood estimation (MLE) optimizer is used.

% Avoid exposure bias
% It attracts many researchers to solve this problem. 
Generative Adversarial Networks (GAN) are one of the directions to solve the exposure bias problem.
%.                  v 'alternately train' is a bit weird wording. TODO (DNGros) propose alternative
The main idea is to alternately train between a discriminator to distinguish real samples from generated samples and the generator to improve its generated samples against the discriminator.
Along this direction, there have been many studies.
% Criticism 1
Nevertheless, there are increasing criticisms \cite{Caccia2020Language,DBLP:conf/naacl/TevetHSB19,DBLP:journals/corr/abs-1806-04936} of text GANs showing that GAN generated text is substantially worse than the text generated by MLE.
Especially, \citet{Caccia2020Language} find that MLE has a better quality-diversity trade-off when using the temperature sweep method for evaluation.
% Criticism 2
% "In the meantime" is seems like potentially weird wording. Do you mean "Concurrently,"? Or "More recently,"? 
More recently, large generative pre-trained language models have greatly improved the quality of MLE generations \cite{Radford2018ImprovingLU, radford2019language}, which further increases the gap of performance between MLE and text GANs.

In this work, we investigate whether large pre-trained language models can improve GANs in text generation.
We propose TextGAIL, a generative adversarial training framework that leverages guidance from the large-scale pre-trained language models RoBERTa \cite{ DBLP:journals/corr/abs-1907-11692} and GPT-2 \cite{radford2019language}.
We find that it does not work by simply combining the previous adversarial approaches with large pre-trained language models due to the high variance in gradients and the architecture limitations.

To reduce variance and improve performance, we apply generative imitation learning (GAIL) \cite{DBLP:conf/nips/HoE16} and proximal policy optimization (PPO) \cite{DBLP:journals/corr/SchulmanWDRK17} for the optimization.
We also introduce contrastive discriminator to better serve the conditional generation tasks.

For a fair comparison, we adopt temperature sweep approach \cite{Caccia2020Language} to evaluate the quality-diversity trade-off.
Previous text GANs often only perform experiment on unconditional generation tasks: COCO and EMNLP2017 News.
%                         v What exactly do you mean by "limited usage scenarios"?
We extend the experiments to conditional generation tasks, as more practical applications. Specifically, we experiment our model on CommonGEN and ROCStories.
% They can more reflect the performance of a model, and we can apply beam search instead of sampling to have deterministic outputs.

We make several contributions:
% Contribution 1
(1) We propose a generative adversarial imitation learning framework TextGAIL, which leverages large pre-trained language models.
% Contribution 2
(2) We conduct extensive evaluations to show TextGAIL achieves better quality and diversity compared to an MLE fine-tuned baseline.
% Contribution 3
(3) We show that large pre-trained language models can help the discriminator to provide useful rewards during the adversarial training process.

% Experimental results show that TextGAIL has better a quality-diversity trade-off compared to MLE using the temperature sweep evaluation metric on a variety of text generation tasks.

\section{Related Work}

Exposure bias is often considered to attribute to low-quality text generations by having generic and repetitive sentences \cite{DBLP:journals/corr/abs-1908-04319,DBLP:journals/corr/abs-1904-09751}.
Even after the emergence of large-scale pretrained language model GPT-2, this problem is still prevalent \cite{DBLP:journals/corr/abs-1908-04319}. 
Many works tried to use Generative Adversarial Networks (GAN) \cite{DBLP:journals/corr/GoodfellowPMXWOCB14} to eliminate the exposure bias problem caused by MLE \cite{DBLP:journals/corr/RanzatoCAZ15,DBLP:journals/corr/abs-1908-04319}.

SeqGAN \cite{DBLP:conf/aaai/YuZWY17} is the very first paper that adopts the adversarial training idea in text generation. 
As text sequence is discrete, SeqGAN applies REINFORCE \cite{DBLP:journals/ml/Williams92}, which is a policy gradient algorithm, to train the generator with a reward defined by the discriminator's prediction on the generated sample.
% Gumbel Softmax
% The alternative is to apply Gumbel Softmax to relax discrete distribution to have a continuous approximation, which enables the gradient back-propagation from the discriminator to the generator \cite{DBLP:journals/corr/KusnerH16}.
% However, these two methods result in a high variance of gradients and are highly unstable during training.
However, it suffers from a high variance of gradients.
% Language model
There are many other text GANs \cite{DBLP:journals/corr/abs-1908-07195,DBLP:journals/corr/CheLZHLSB17,DBLP:conf/aaai/GuoLCZYW18,DBLP:conf/nips/LinLHSZ17,DBLP:conf/iclr/NieNP19,Zhou2020Self-Adversarial}.
However, as \citet{Caccia2020Language} has shown, many of them are worse than their MLE counterparts when evaluated in the quality-diversity trade-off setting. 
This is because many text GANs assume MLE-based models keep the softmax temperature to be $1.0$ when sampling, but MLE-based models actually perform better with a lower temperature.
Consequently, after using temperature sweep, MLE-based method has a better quality-diversity trade-off curve than many text GANs.

In another line of works, large scale pre-training \cite{radford2019language} has shown significant improvement in text generation. 
Large pre-trained models such as GPT-2 can be fine-tuned with MLE on a specific task to achieve much better performance than the models without pre-training.
Some papers even claim human-level text generation quality \cite{DBLP:journals/corr/abs-2001-09977}. 
It is interesting to explore whether text GANs can be combined with large pre-trained language models to improve performance further. 
% However, since some previous text GANs involve architecture changes such as LeakGAN \cite{DBLP:conf/aaai/GuoLCZYW18} and RelGAN \cite{DBLP:conf/iclr/NieNP19}, it is hard to directly apply them on the Transformer-based \cite{DBLP:conf/nips/VaswaniSPUJGKP17} models. 
In this work, we propose an new imitation learning framework to combine pre-training models with GANs.

\section{TextGAIL}

\begin{figure*}[ht]
    \centering
    \includegraphics[width=0.95\textwidth]{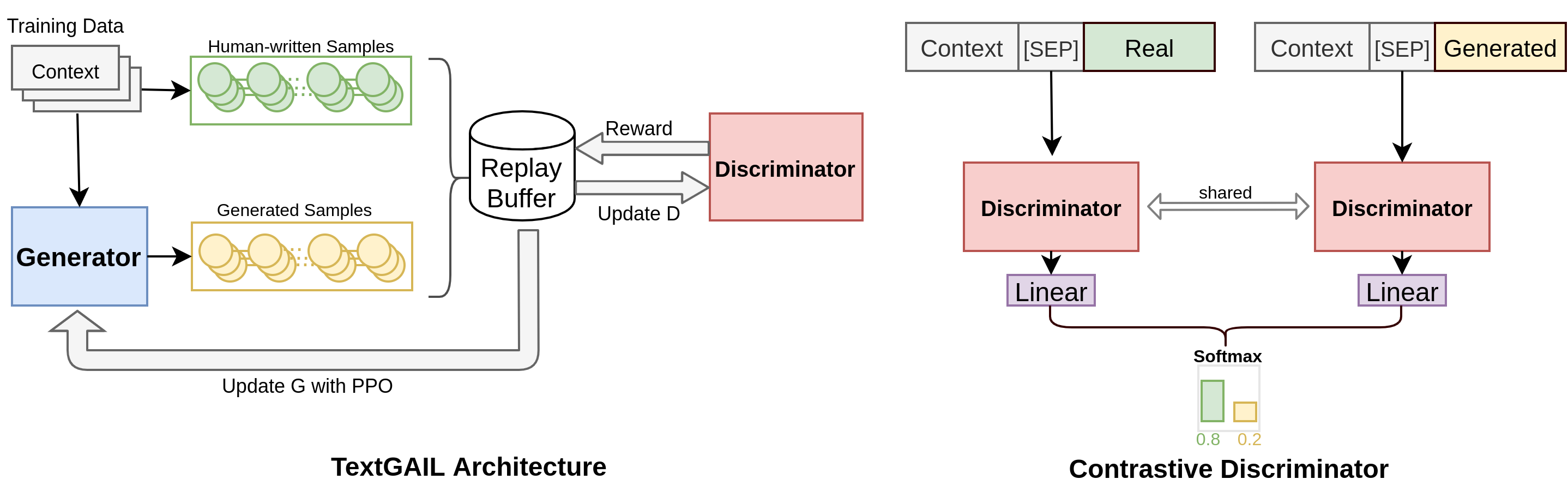}
    \caption{\textbf{Left}: Overall architecture of TextGAIL. 
    \textbf{Right}: The contrastive discriminator..
    }
    \label{fig:model}
\end{figure*}

% \begin{figure}[t]
%     \centering
%     \includegraphics[width=0.45\textwidth]{images/example.png}
%     \caption{An example of story ending generation in TextGAIL. D is the discriminator. G is the generator.}
%     \label{fig:example}
% \end{figure}

In this section, we first give an overview of the generative adversarial imitation learning framework.
Then we explain the discriminator and the generator in details. 
% They are augmented with the large pre-trained language models, RoBERTa and GPT-2 respectively to receive knowledge from large scale pre-training.
% We further combine efficient use of human demonstrations to stabilize the training.
In the end, we summarize the entire training process. 
We show the overall architecture in Figure~\ref{fig:model}.

\subsection{Generative Adversarial Imitation Learning }

We extend the generative adversarial imitation learning (GAIL) \cite{DBLP:conf/nips/HoE16} to text generation.
The framework consists of a generator $G_\theta$ and a discriminator $D_\phi$, which are parameterized with $\theta$ and $\phi$, respectively.
The goal of the generator is to output sequences similar to human written sequences.
Meanwhile, the discriminator needs to distinguish the real sequences from the generated sequences, and provide a single sparse reward for each generated sequence.

Here, we replace the state $s$ in GAIL with the text generation prompt $x$, and the corresponding action $a$ with the target sequence $y$.  
Note that in the unconditional generation setting, $x$ can be the start token.
$y$ can either be given from ground truth in the dataset as real data or sampled from the generator $G_\theta$ as fake data.
GAIL finds a saddle point where together the generator and discriminator satisfy the following objective function:
\begin{equation}
    \underset{G_\theta}{\text{min}} \, \underset{D_\phi}{\text{max}} \:  
     \: \mathbb{E}_{p_{\mathrm{real}}} [ D_\phi(x, y)] \: + 
     \mathbb{E}_{G_\theta}[ 1 - D_\phi(x, G_\theta(x))]
    % +   \text{KL}(q|p_\theta)
\label{eq:gail}
\end{equation}

However, in text generation, the action space, which is the vocabulary size, is often vary large.
The original GAIL has difficulty to remain stable with such a large action space \cite{DBLP:journals/corr/abs-1909-01387}.
We introduce an imitation replay method inspired by the recent imitation learning algorithms \cite{DBLP:journals/corr/abs-1909-01387, DBLP:conf/iclr/ReddyDL20} to stabilize the training.

We fill the experience replay buffer with a ratio $\lambda$ of ground truth sequences when training the generator. 
Those ground truth sequences are treated the same as the generated sequences in the replay buffer.
We set the reward (without normalization) to be a constant for the ground truth sequences.
% Therefore, 
This approach is theoretically similar to mixing supervised MLE loss during the training, but in practice, it is much more efficient and easier to implement.

% % How to solve it
% Fortunately, the recent work of DeepMind's Recurrent Replay Distributed DQN from Demonstrations (R2D3) \cite{DBLP:journals/corr/abs-1909-01387} introduced another imitation learning method that efficiently uses human demonstrations to solve hard exploration problems in RL. 
% We apply this method to the TextGAIL framework.

% how to do it
% During the training of the generator, we introduce a hyper-parameter $p$ to control the ratio between the sampling from human demonstrations and generated sequences. 
% The human demonstrations are treated the same as generated sequences to be put into the replay buffer, but we force the reward for human demonstrations to be a constant.
% By fixing the human demonstrations' advantage to be a constant, it stabilizes the training to update the generator distribution to be close to the real data distribution.

\subsection{Contrastive Discriminator}

% What does the discriminator do?
The discriminator aims to distinguish between the real and generated samples.
% The problem
Standard discriminator utilizes logistic loss (sigmoid), but this loss saturates quickly after the model learns the difference between the real and the generated samples.
We modify the discriminator to be a contrastive discriminator, which estimates the relative realness between generated sequences and real sequences. 
In other words, we let the discriminator estimate how much a real sequence is more realistic than a generated sequence.
This can especially help conditional generation tasks.
% The idea is similar to \citet{DBLP:conf/iclr/Jolicoeur-Martineau19}.
Here, we perform the prediction by utilizing softmax cross-entropy instead of logistic loss.

Instead of $D_\phi(x, y)$, the discriminator now takes a real sequence and its paired generated sequence as inputs, denoted as $D_\phi(\langle x, y_\mathrm{r} \rangle, \langle x, y_\mathrm{g} \rangle)$.
The discriminator outputs a score to represent how good is the generated sequence $y_\mathrm{g}$ compared with the real sequence.
% We use RoBERTa as the base classifier for the discriminator.
Then we optimize it with the following objective function.
\begin{align}
    h_\mathrm{r} &= \mathrm{Discriminator} (\langle x, y_\mathrm{r} \rangle) \\
    h_\mathrm{g} &= \mathrm{Discriminator} (\langle x, y_\mathrm{g} \rangle) \\
    p_r, p_g & = \mathrm{softmax}(W_t [h_\mathrm{r};h_\mathrm{g}])
    \label{eq:discriminator}
\end{align}
where $W_t$ is the trainable weight to project the output embedding to a scalar logit. $y_\mathrm{r}$ is the real sequence, and $y_\mathrm{g}$ is the generated sequence.
We can optimize the discriminator with cross-entropy loss to maximize the probability $p_r$ for the real sequence.
The probability prediction $p_g$ for the generated sequence will be used as the reward signal to train the generator.

\subsection{Proximally Optimized Generator}

For the generator, we begin by defining the probability of a text sequence as the joint probability of all the tokens: 
\begin{equation}
    G_\theta(y_{1:T} | x) = \prod_{t=0}^T G_\theta(y_t | y_{<t}, x)
\end{equation}
where $y_{1:T}$ is a text sequence. 
$T$ is the sequence length, and $y_t$ is the word at the time step $t$.
%                               ^ do you mean $y_t$? It doesn't look like $w_t$ is used in the equation?
% Sampling
We sample from this distribution to acquire the generated sequences.
Then we maximize the expected reward with policy gradient:

\begin{equation}
    \mathbb{E}_{y \sim G_\theta}[\nabla_\theta \, \mathrm{log} \, G_\theta(x) \hat{R}_{y}]
    \label{eq:reward}
\end{equation}
where $\hat{R}_{y}$ is the advantage term that controls the update (which is the normalized reward here). 
Directly optimizing this objective suffers from high variance of gradients, because the $D_\phi$ is not stationary during adversarial training. 

% What is new in our method
% why we replace TRPO
% Our Modification
As a solution to reduce high variance, the original GAIL employs trust region policy optimization (TRPO) \cite{DBLP:conf/icml/SchulmanLAJM15}, as it is crucial to ensure that $G_{\theta_{i+1}}$ does not move too far away from $G_{\theta_i}$.
However, TRPO needs to compute natural gradient which is computationally expensive.
We replace it with a more recent and stable method, proximal policy optimization (PPO) \cite{DBLP:journals/corr/SchulmanWDRK17}.
% Why
Compared to TRPO, PPO is easier to implement and generalize. 
PPO has better sample complexity in practice as well.

PPO applies importance sampling by the likelihood ratio between the current and old policy for $y \sim G_{\theta_{\text{old}}} (\cdot | x)$:
\begin{equation}
    r(\theta)  = \frac{G_{\theta}(y_{1:T}  \, | \,  x)}{G_{\theta_{\text{old}}}(y_{1:T}  \, | \, x)}
\end{equation}
Then it maximizes the expected reward by optimizing the following surrogate:
\begin{equation}
    L_\text{G}(\theta) = - \text{min}
    \begin{cases}
        r (\theta) \, \hat{R}_{y} \\
        \text{clip} \, (r (\theta), 1 - \epsilon, 1 + \epsilon) \, \hat{R}_{y} \\
    \end{cases}
    \label{eq:generator}
\end{equation}
This surrogate serves the same purpose as TRPO to have a trust region constraint on the gradient update. 
It will prevent the generator from moving too far away from the pre-trained language model.

% In addition, we add the KL divergence between the generator and another pre-trained language model in replacement of the traditional maximum entropy regularizer. 
% This KL term is part of the contributions of the paper, correct? Not just an extension of the prior explanation of PPO/GAIL? Perhaps 
% consider clarifying whether this is part of the contribution or not. "For TextGAIL, we add..." or some other wording to clarify
% what is the source of this KL-term idea.
% The final generator objective can be written as the following:
% \begin{equation}
%     L_G(\theta) = - \underset{y \sim p_\theta(\cdot|x)}{\mathbb{E}} [ L_\text{policy} (\theta)  
%     % + \beta \, \text{KL} (q | \, p_\theta) ]
%     \label{eq:generator}
% \end{equation}
% $q$ is the pre-trained language model.
% This new term stimulates more exploration while not affecting the original language model distribution.

% \subsection{Efficient Use of Human Demonstrations}

\subsection{TextGAIL Training Process}

% % Now it is the time to combine everything together.
% Finally, we combine everything and present the algorithm of TextGAIL.
We warm-up the generator by training with a part of the training set using MLE as its loss function.
We alternately train the discriminator and the generator to optimize Equation~\ref{eq:gail}.
A replay buffer stores temporarily generated outputs and human-written sequences.
% After the replay buffer is full, the discriminator would assign rewards to all the context-target pairs in the replay buffer.
Next, we normalize the rewards in the buffer with running statistics to reduce variance.
We update the generator $G_\theta$ with PPO using the replay buffer.
In the meantime, we update the discriminator $D_\phi$ with the real and generated pairs.
We repeat until the training stops.
The summary of the algorithm is illustrated below.

\begin{center}
\begin{minipage}{1.0\linewidth}
\begin{algorithm}[H]
\caption{TextGAIL}
% \small
\begin{algorithmic}[1]
\STATE \textbf{Initialize}: Collect human-written sequences \\ 
                            \qquad \qquad \, Warm-up the generator $G_\theta$ \\
                            \qquad \qquad \, Replay Buffer $B$
\FOR{i = $1,2,3, \dots$}
    \STATE Sample $p$ proportion of human-written sequences $y$
    \STATE Sample $1 - p$ proportion of generator outputs $y \sim G(\cdot | x)$
    \STATE Put all sampled $(x, y)$ pairs into $B$
    \STATE Collect rewards using discriminator $D_\phi$ for all $(x, y) \in B$
    \STATE Normalize all the rewards to get $\hat{R}$
    \STATE Replace rewards for human-written sequences with a constant
    \STATE Update the discriminator $\phi$ with Eq.~\ref{eq:discriminator}
    \STATE Update the generator $\theta$ using the PPO with Eq.~\ref{eq:generator}
    \STATE Clear Buffer $B$
\ENDFOR
\end{algorithmic}
\label{algorithm}
\end{algorithm}
\end{minipage}    
\end{center}

\section{Experimental Settings}
% [WE NEED A PARAGRAPH HERE TO SUMMARIZE THE SUBSECTIONS]
% How to prove adversarial training can help
%To test the effectiveness of adversarial training, we first train a discriminator based on a MLE loss using the generator's outputs.
%This experiment reveals how the common sense knowledge is used by  

We will describe the datasets, implementation details and automatic evaluation metrics in this section.

\subsection{Datasets}

We apply TextGAIL on both unconditional and conditional text generation tasks.
We use COCO Image Captions and EMNLP2017 News (the same datasets used in previous text GANs) \cite{DBLP:conf/aaai/YuZWY17} as unconditional generation tasks. 
We use CommonGEN \cite{DBLP:journals/corr/abs-1911-03705}, and ROCStories (story ending generation) \cite{DBLP:journals/corr/MostafazadehCHP16} as conditional generation tasks. 
The details of statistics of each dataset are in Table~\ref{tab:datasets}. 
% ^ link to proper appendix section?
% TODO
% Table~\ref{tab:datasets} shows the statistics of the datasets.
% COCO is about generating an image caption.

\begin{table}[h]
    \centering
    \resizebox{0.47\textwidth}{!}{
        \begin{tabular}{lcccc}
            \toprule
                                  & COCO     & EMNLP2017 WMT & CommonGEN    & ROCStories   \\
            \midrule
            Condition generation   & $\times$ & $\times$      & $\checkmark$ & $\checkmark$ \\
            \midrule
            Vocabulary size        & 4,681    & 12,972        & 12,941       & 20,937       \\
            \midrule
            Average length         & 11.36    & 27.03         & 10.97        & 10.59        \\
            \midrule
            \# of Training Samples & 10,000   & 278,586       & 39,069       & 98,161       \\
            \bottomrule
        \end{tabular}
    }
    \caption{Datasets Statistics.
    }
    \label{tab:datasets}
\end{table}

\subsection{Implementation Details}

TextGAIL takes advantage of large pre-trained language models. 
In particular, the generator uses the GPT-2 base (117M parameters) model, while the discriminator uses the RoBERTa-base (125M parameters) model.
The human demonstrations mix ratio $p$ is set to $0.3$ at the start of the training and linearly decay afterward.
The constant reward for human demonstrations is set to $2.0$.
When generating outputs, we apply the recent nucleus sampling method
\cite{DBLP:journals/corr/abs-1904-09751} for decoding to avoid low probability words being sampled.
We stop the training when the perplexity stops decreasing for both MLE and TextGAIL.
The details of hyper-parameters are in the Appendix.
% ^ what top-p threshold is used during the nucleus sampling?
% Also, is this the same as was used in Caccia et al? This seems like a potential extra variable in the diversity measure which
% a reviewer might question?

\section{Baselines}

Since previous text GANs are mainly for unconditional tasks, we only show their performance on unconditional tasks.
% Also, there are various criticisms \cite{Caccia2020Language,DBLP:conf/naacl/TevetHSB19,DBLP:journals/corr/abs-1806-04936} that point out text GANs are worse than MLE-based models.
For conditional generation tasks, we compare TextGAIL with GPT-2 fine-tuned on training dataset with a MLE loss.
For a fair comparison between MLE models and TextGAIL models, we stop the training when TextGAIL reaches the perplexity of MLE baselines.
% This is because sometimes over-fitting with worse perplexities can have better performance on other metrics such as BLEU.

\subsection{Evaluation Metrics}
We measure model's quality and diversity from a range of temperatures between $0.1$ to $1.0$.
This temperature sweep method ensures fair comparisons as described by \citet{Caccia2020Language}

% Quality
For the quality metric, we use the n-gram matching metric BLEU.
When using BLEU for unconditional generation tasks, the entire training corpus is used as references for BLEU \cite{DBLP:conf/aaai/YuZWY17}.
Since BLEU has its limitations, we further conduct human evaluations to measure models' generation quality.
We also compare perplexity under different temperatures for more comprehensive comparisons.

% Diversity
For the diversity metric, we use Self-BLEU \cite{DBLP:conf/aaai/YuZWY17} on unconditional generation tasks, and Distinct-n \cite{DBLP:conf/naacl/LiGBGD16} on conditional generation tasks. 
Self-BLEU evaluates how one generated sentence resembles the rest in a set of generated samples.
Distinct-n is the number of distinct n-grams divided by the total number of n-grams in the test set.
When decoding with beam search, we use Seq-Rep-n \cite{DBLP:journals/corr/abs-1908-04319} to measure sequence-level repetition inside a sentence. Seq-Rep-n is the portion of duplicate n-grams in a sequence.
% Some Explanation of the metrics

\begin{figure}[h]
    \centering
    \begin{subfigure}[b]{0.4\textwidth}
        \centering
        \includegraphics[width=\textwidth]{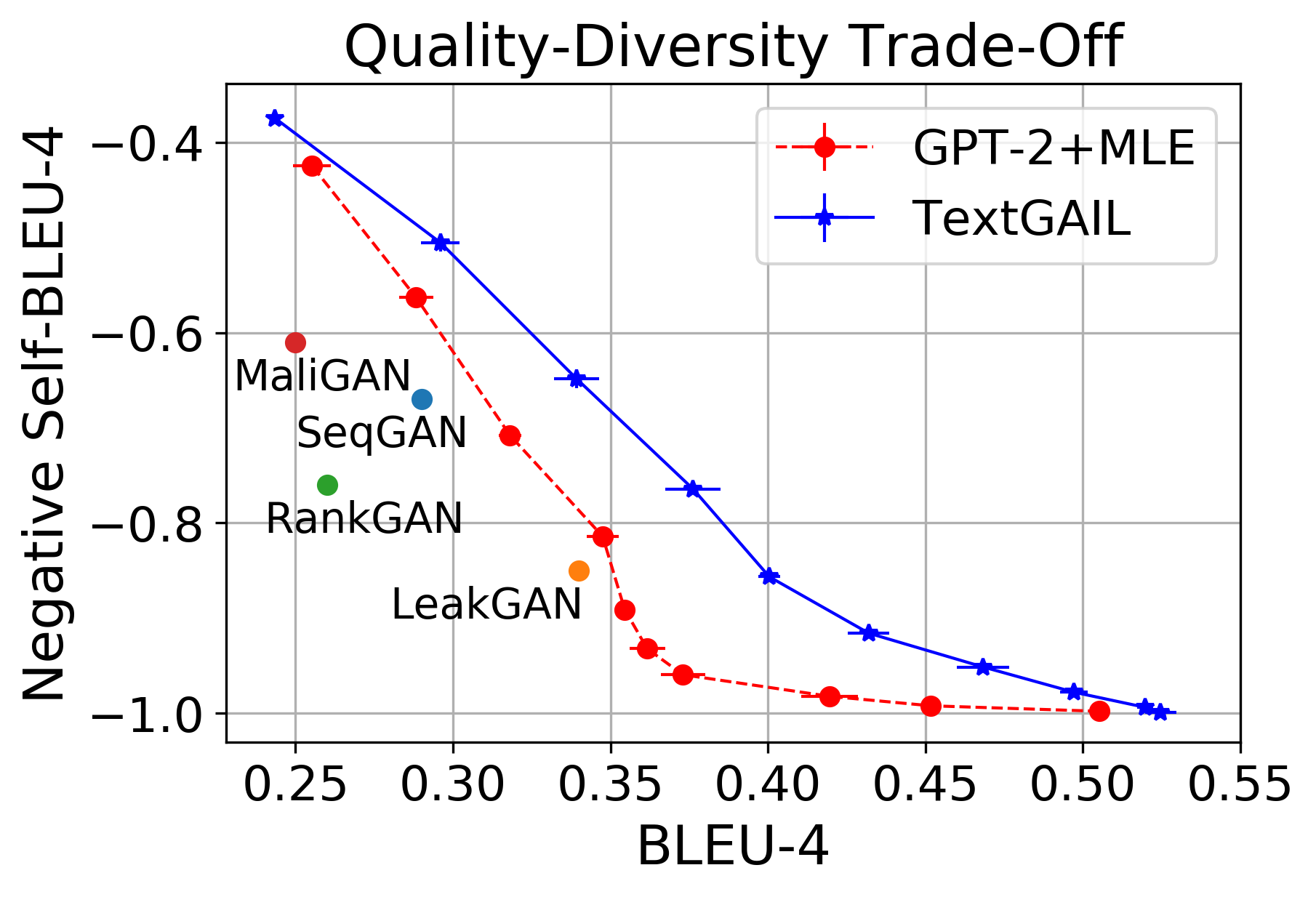}
        \caption{COCO Captions}
    \end{subfigure}
    \hfill
    \begin{subfigure}[b]{0.4\textwidth}
        \centering
        \includegraphics[width=\textwidth]{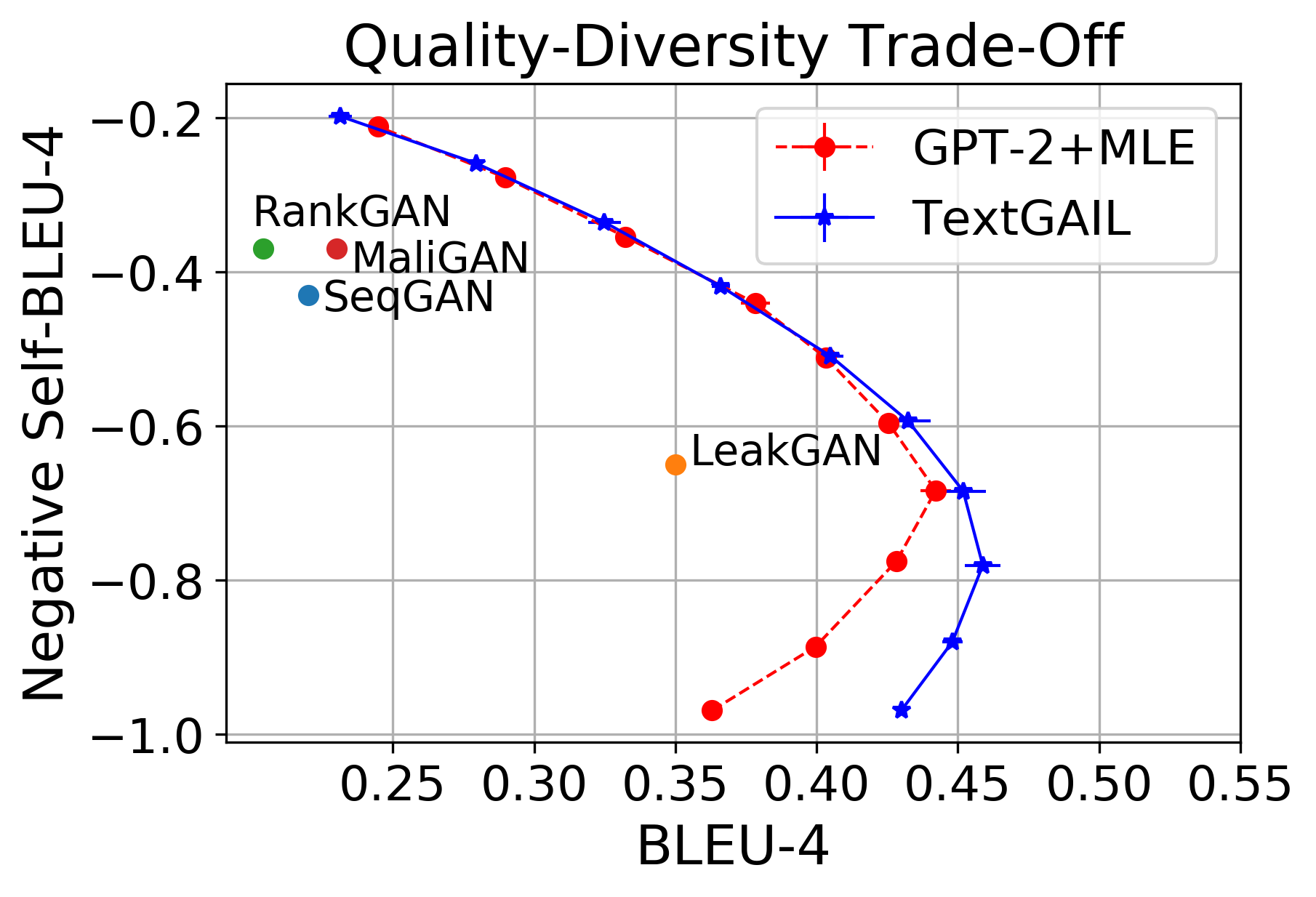}
        \caption{EMNLP2017 News}
    \end{subfigure}
    \caption{
   Quality-diversity trade-off on unconditional tasks.
   The curve closer to the top right corner has better performance.
   Error bars are from three random seed runs.
    }
    \label{fig:unconditional}
\end{figure}

\begin{figure}[h]
    \centering
    % \quad
    \begin{subfigure}[b]{0.4\textwidth}
        \centering
        \includegraphics[width=0.9\textwidth]{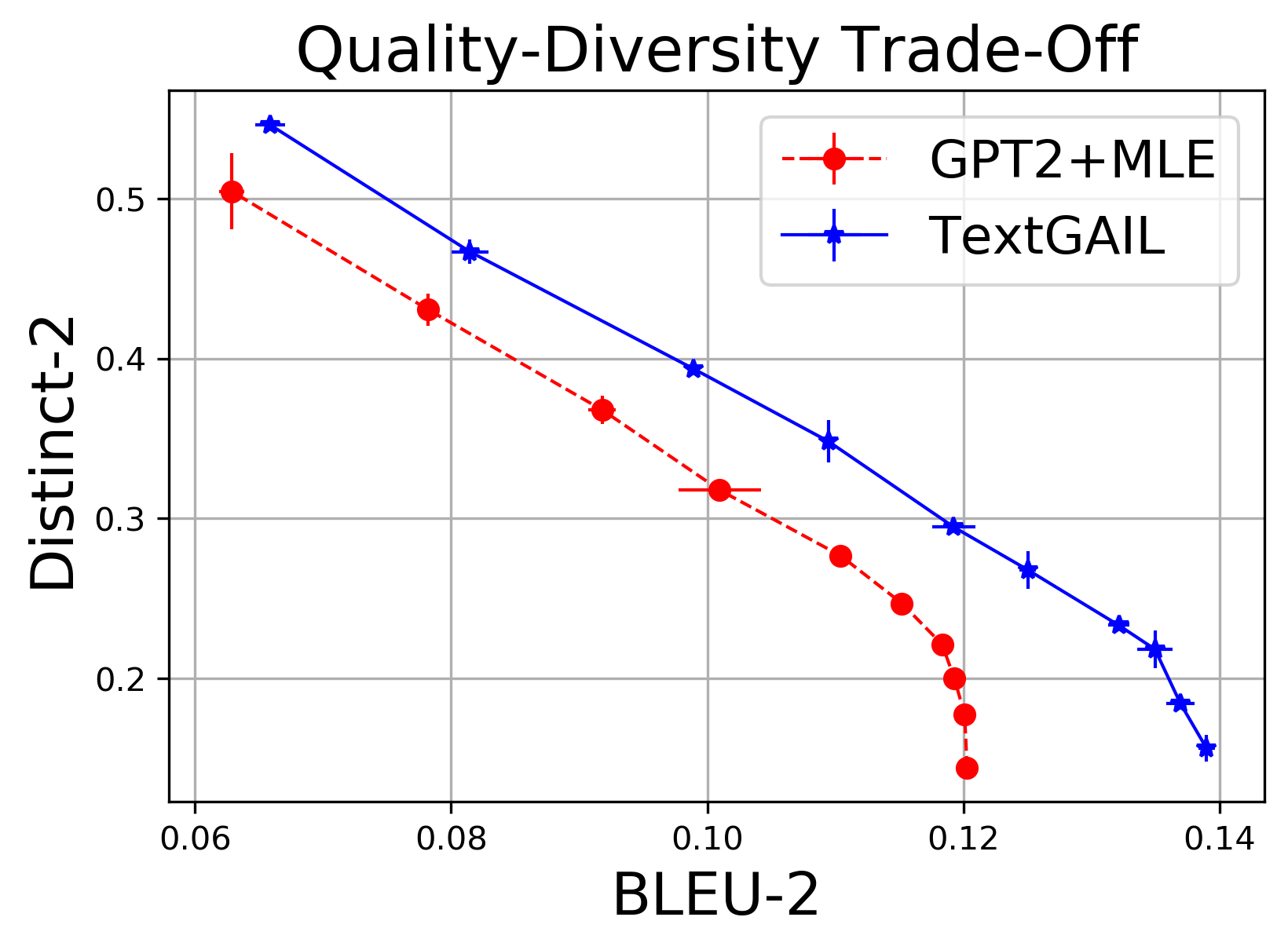}
        \caption{CommonGEN}
    \end{subfigure}
    \hfill
    \begin{subfigure}[b]{0.4\textwidth}
        \centering
        \includegraphics[width=0.9\textwidth]{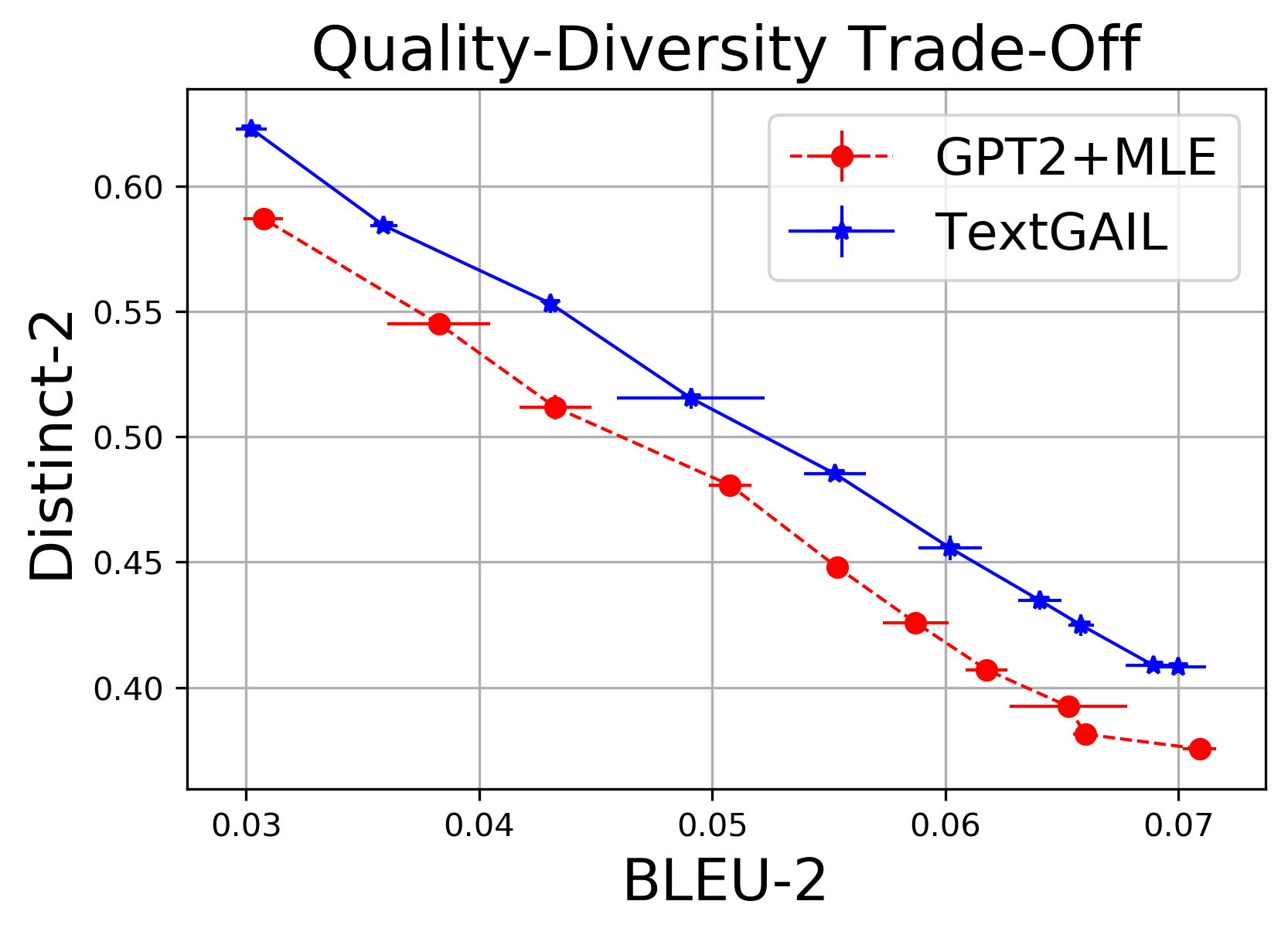}
        \caption{ROCStories}
    \end{subfigure}
    % What is Distinct-2? Is it distinct bigrams? (perhaps specify?). Why is Distinct-2 used for the contextual generation tasks rather
    % than the Negative Self-BLEU-4 as used in (a) (b)
    % TODO
    \caption{
   Quality-diversity trade-off on conditional tasks.
   The curve closer to the top right corner has better performance.
   Note that there are no results from the  most previous GANs, as they are not designed for conditional tasks.
   Error bars are from three random seed runs.
    }
    \label{fig:conditional}
\end{figure}

\section{Results and Analysis}

\subsection{Automatic Evaluation Results}

We first test if TextGAIL performs better than a model that fine-tunes GPT-2 on the training dataset with an MLE loss. 
As suggested by \citet{Caccia2020Language}, temperature sweep can reflect the quality-diversity trade-off, which enables fair comparisons between two models.
We sweep the softmax temperature from $0.1$ to $1.0$ to observe how the models behave accordingly.
For unconditional generation tasks, we report BLEU vs. Self-BLEU as the quality-diversity metric (we use negative Self-BLEU for better visualization).
% Consider reporting somewhere what form of smoothing (if any) is used on the BLEU, or which implementation of BLEU is used
For conditional generation tasks, we report BLEU vs. Distinct as the quality-diversity metric.
% We also report how perplexity changes when temperature varies.
The blue lines are TextGAIL, and the red lines are GPT-2 fine-tuned with MLE.

% All experimental results are shown in Figure~\ref{fig:metrics}. 
% In the perplexity-temperature figures, the two models' curves are close to each other.
% One reason that the two lines are close is that we stop the training when the perplexity stops decreasing on the validation set.
% This suggests TextGAIL can achieve similar perplexity scores as GPT-2+MLE. When the temperature is low, TextGAIL is slightly better than MLE in perplexity.

For the unconditional tasks, the results are shown in Figure~\ref{fig:unconditional}.
TextGAIL and GPT-2+MLE is much better than the previous text GANs.
For COCO Captions, we can observe that TextGAIL achieves great improvement over MLE.
However, for EMNLP2017 News, the difference is not significant.
We suspect that the reason is because of the text length in the dataset.
EMNLP2017 News has much longer text length than COCO Captions.
This may lead to worse reward signals from the discriminator when training the generator.

In Figure~\ref{fig:conditional}, we show the results of conditional tasks.
Since previous text GANs are not designed for conditional generation tasks, we are unable to report them here.
From the figure, we can observe more consistent improvement of TextGAIL compared to the improvement in unconditional tasks.
While in unconditional generation tasks, without the context given, it is even hard for humans to classify if the text is real or generated.

\begin{table}[h]
        \begin{center}
            \resizebox{0.45\textwidth}{!}{
                \begin{tabular}{lcccc}
                    \toprule
                                           & \multicolumn{2}{c}{CommonGEN} & \multicolumn{2}{c}{ROCStories}                            \\
                                           & MLE                            & TextGAIL                       & MLE    & TextGAIL        \\
                    \midrule
                    BLEU-2 $\uparrow$      & 15.74 $\pm$ 0.04                          & \textbf{16.21} $\pm$ 0.17                 & 7.44 $\pm$ 0.13 & \textbf{7.77} $\pm$ 0.27 \\
                    \midrule
                    Distinct-2 $\uparrow$  & 8.66 $\pm$ 0.05                         & \textbf{10.00} $\pm$ 0.08              & 41.05 $\pm$ 2.39 & \textbf{46.83} $\pm$ 3.83 \\
                    \midrule
                    Seq-Rep-2 $\downarrow$ & 65.85$\pm$ 2.30                        & \textbf{60.78} $\pm$ 2.41              & 76.92 $\pm$ 2.18 & \textbf{74.71} $\pm$ 2.07 \\
                    \bottomrule
                \end{tabular}
            }
        \end{center}
    \caption{Beam search results on the conditional tasks. Since beam search results are deterministic, it is not applicable to the unconditional tasks.}
    \label{tab:beam_search}
\end{table}

% Beam search results
One advantage of testing with conditional tasks is that we can apply deterministic decoding methods such as beam search rather than stochastic decoding used in unconditional generation tasks.
This can provide more reliable interpretation of our model.

We perform beam search with a beam size of four on the two conditional generation tasks.
We run the experiments with three different random seeds.
The results on beam search generations are shown in Table~\ref{tab:beam_search}.
We observe that TextGAIL performs better than MLE-based method in both quality and diversity metrics.
When we examine the generated text, we find that MLE produces many repetitions, which is possibly caused by text degeneration with exposure bias \cite{DBLP:journals/corr/abs-1908-04319}.
As TextGAIL mitigates exposure bias, both the quality metric BLEU-2 and the diversity metric Distinct-2 improves over the baseline MLE method.

We suspect the main contribution of improvement is due to the discriminator being more effective in providing useful reward signals in conditional generation tasks.
As in CommenGEN and ROCStories, the contrastive discriminator can better classify the realness of the generations conditioned to the context by comparing the real and generated sentences.
In Section 6, we will conduct additional experiments to interpret what the discriminator has learned.

% ^ is this a qualitative claim? If not consider stating what metric backs this up. "... we observe that there is less repetition without
%        sacrificing quality as indicated by higher Distinct-2 at equivalent BLEU scores." (not sure if this is factual. Also, as mentioned
%        unsure what this Distinct-2 that is used on the graph is). If the claim is qualitative (which is my impression from following up on "When we examine the generations..."), consider ways of adding a empirical
%        measure of repeats.
\subsection{Human Evaluation Results}

\begin{table}[h]
    \centering
    \footnotesize
    \resizebox{0.4\textwidth}{!}{
        \begin{tabular}{lccc}
            \toprule
                       & \multicolumn{3}{c}{TextGAIL vs MLE}                    \\ \cmidrule{2-4}
                       & Win                                 & Lose    & Tie    \\
            \midrule
            COCO       & 31.2\%                              & 27.6\%  & 41.2\% \\
            \midrule
            EMNLP2017  & 44.2 \%                             & 44.2\%  & 11.6\% \\
            \midrule
            CommonGEN  & 59.6\%*                             & 34.6\%* & 5.8\%  \\
            \midrule
            ROCStories & 60.5\%*                             & 23.4\%* & 16.1\% \\
            \bottomrule
        \end{tabular}
    }
    \caption{Human evaluation results. *denotes statistical significance (binomial test, $p < 0.05$)}
    % What kind of statistical test is used? What is the null hypothesis? Consider mentioning here or in the text..
    % There is nuance here since have win/loss/ties which makes it less obvious what is used compared to simple measurements.
    \label{tab:human_eval}
\end{table}

\begin{table*}[h]
    \begin{center}
        \resizebox{1.\textwidth}{!}{
            \begin{tabular}{lll}
                \toprule % <-- Toprule here
                \textbf{CommonGEN}              & Example 1                                                                              & Example 2 \\
                \midrule
                \textbf{Context:}      & \multicolumn{1}{p{8cm}}{\raggedright field look stand}
                                       &
                \multicolumn{1}{p{7cm}}{\raggedright ocean surf surfer}
                \\
                \midrule
                \textbf{Ground Truth:} & \multicolumn{1}{p{8cm}}{\raggedright I stood and looked across the field, peacefully.}
                                       &
                \multicolumn{1}{p{7cm}}{\raggedright A surfer surfing in the ocean.}
                \\
                \midrule
                \textbf{MLE:}          & \multicolumn{1}{p{8cm}}{\raggedright
                    (1) looks at the ground at the end of the driveway                                                                      \\
                    (2) looks at the wall and the wall stands                                                                               \\
                    (3) i stand on a bench in front of the field with a smile on my lips
                }                      & \multicolumn{1}{p{7cm}}{\raggedright
                (1) surfer and surfers walk the beach at the coast                                                                          \\
                (2) surfer in the surf on the coast                                                                                         \\
                (3) surfer in the ocean.
                }                                                                                                                           \\
                \midrule
                \textbf{TextGAIL:}     & \multicolumn{1}{p{8cm}}{\raggedright
                    (1) field looks like a soccer field with a few soccer players standing                                                  \\
                    (2) a man stands in the middle of the field looking at the scoreboard                                                   \\
                    (3) a small group of people stand in the field looking at a city
                }                      & \multicolumn{1}{p{7cm}}{\raggedright
                (1) the surfers wave their surfboards on the beach                                                                          \\
                (2) Two surfers are surfing in the ocean and one is looking to the horizon.                                                 \\
                (3) a surf diver watches as a group of dolphins swim in the ocean
                }                                                                                                                           \\
                \bottomrule % <-- Bottomrule here
            \end{tabular}
        }
    \end{center}
    \caption{Examples of MLE and TextGAIL on CommonGEN. TextGAIL follow the instruction by using the three given three words in the generations. The results are longer and more diverse than MLE.
    }
    \label{tab:commongen_generations}
\end{table*}

\begin{table*}[h]
    \begin{center}
        \resizebox{1.\textwidth}{!}{
            \begin{tabular}{lll}
                \toprule % <-- Toprule here
                \textbf{ROCStories}             & Example 1                                                                                                                                                                                                                                                         & Example 2 \\
                \midrule
                \textbf{Context:}      & \multicolumn{1}{p{8cm}}{\raggedright I wanted to buy a video game console. I asked my parents, and they came up with an idea. They said if I did my chores, I would be given money to save. I did my chores without being asked every week for a whole summer.  }
                                       &
                \multicolumn{1}{p{7cm}}{\raggedright Ben went to the DMV to get his License. The instructor gave Ben a passing grade at the end. Excited, Ben calls up his father to tell him the good news. Ben father never picked up, he died in a car accident that day.  }
                \\
                \midrule
                \textbf{Ground Truth:} & \multicolumn{1}{p{8cm}}{\raggedright My parents gave me enough money to buy the console.}
                                       &
                \multicolumn{1}{p{7cm}}{\raggedright Ben was devastated.}
                \\
                \midrule
                \textbf{MLE:}          & \multicolumn{1}{p{8cm}}{\raggedright
                    (1) Now I have the video game console I asked for.                                                                                                                                                                                                                                                 \\
                    (2) It was an awesome idea.                                                                                                                                                                                                                                                                        \\
                    (3) The next week, I had to buy a new gaming console!
                }                      & \multicolumn{1}{p{7cm}}{\raggedright
                (1) Ben was happy to learn his lessons about being smart.                                                                                                                                                                                                                                              \\
                (2) Ben's father is now very sad, and he has a job to do.                                                                                                                                                                                                                                              \\
                (3) Ben was happy that his dad was alive.
                }                                                                                                                                                                                                                                                                                                      \\
                \midrule
                \textbf{TextGAIL:}     & \multicolumn{1}{p{8cm}}{\raggedright
                    (1) I bought a PlayStation 4 to play with my parents.                                                                                                                                                                                                                                              \\
                    (2) I was so happy when my parents gave me a Wii U.                                                                                                                                                                                                                                                \\
                    (3) When I got my console, I played my favorite video games.
                }                      & \multicolumn{1}{p{7cm}}{\raggedright
                (1) Ben regrets going to the DMV.                                                                                                                                                                                                                                                                      \\
                (2) Ben mourns the loss of his father but also the passing of a great man.                                                                                                                                                                                                                             \\
                (3) It seems like too much to bear.
                }                                                                                                                                                                                                                                                                                                      \\
                \bottomrule % <-- Bottomrule here
            \end{tabular}
        }
    \end{center}
    \caption{Examples of MLE and TextGAIL on ROCStories. TextGAIL generates better and more reasonable story endings. Also, note that named entities such as "PlayStation 4" and "Wii U" have never appeared in the training set.
    }
    \label{tab:rocstories_generations}
\end{table*}

% Why we do human evaluations
Automatic evaluation metrics have their limitations in measuring overall performance. 
Therefore, we also conduct human evaluations to measure the quality of TextGAIL compared to GPT-2 fine-tuned with MLE.
% Some settings
The MLE model with a low temperature generates large amount of repetitions. We observe the model has less repetition and better quality with nucleus sampling with hyper-parameters top-p $0.9$ and temperature $0.8$. So we use this setting for human evaluation.
For each task, we randomly select 100 samples in test sets, and for each sample, we ask five Amazon Mechanical Turk workers to select which model's result is better to reduce the variance.
In total, we have 500 data points for each task.
For conditional generation tasks, the context is shown to all workers.
The workers are instructed to select the model with better logic and commonsense.
The evaluators can select "Cannot determine" when the two models are similar in quality.

% If possible, would like to see more details. For example how many human annotation were collected? Did you have more than
% one annotator annotate the same example or not? If so could possibly quantitative support the claim about consistency. Also, where
% several stochastic examples compared?
The human evaluation results are shown in Table~\ref{tab:human_eval}.
There is no statistical difference between TextGAIL and MLE on unconditional generation tasks in this pairwise comparison evaluation.
One possible reason is that when comparing two completely different sentences in this unconditional generation setting,
it is difficult for human evaluators to make consistent decisions.
%Since GPT-2 fine-tuned with MLE can already generate high-quality text.
%Thus, we need to experiment more on the conditional generation tasks for better evaluation.
In contrast, TextGAIL significantly outperforms MLE in human evaluation on two conditional generation tasks. Since these tasks expect models to produce similar content with respect to the ground truth. It is easier for human to select the output with better quality.
The results are in agreement with automatic evaluation results.
In the Case Study, we will further analyze the examples .

% This isn't at all needed, but it would have possibly have been nice to also have this data in comparison to the
% human ground truth. Again, this is isn't strictly needed to answer the research question, but would have been nice.

\subsection{Case Study}

We further analyze the generated outputs from the two conditional generation tasks. 
We show some examples in Table~\ref{tab:commongen_generations} and Table~\ref{tab:rocstories_generations}.

% Background info
\textbf{CommonGEN} In the CommonGEN examples, the good target sentences should use the given three words as much as possible. 
% Observation
We can observe that MLE's outputs do not seem to follow that instruction, while TextGAIL is behaving better.
% Explain
This difference suggests that TextGAIL's discriminator might have learned to guide the generator to follow the implied instruction.
% Another observation
Also, we can observe that the MLE's outputs have more repetitions and are less diverse than MLE's outputs. 
This is also partially reflected in the automatic evaluation metrics.
% Explain
These examples also correlates with our intuition that eliminating exposure bias alleviates dull and repetitive outputs \cite{DBLP:journals/corr/abs-1908-04319}.

% Background info
\textbf{ROCStories} For ROCStories, the good story endings should be as reasonable and interesting as possible.
% Observation
We can observe similar patterns in this task.
MLE's generations lack of details and are universal, as the Example 1 MLE (2) appears more than once in other story contexts.
We further find that TextGAIL can generate new named entities such as "PlayStation 4" and "Wii U", which never appeared in the training set. We speculate it might have appeared in GPT-2's pre-trained corpus.
% Explain
This task also provides the evidence that eliminating exposure bias improves generalization of unseen data.
% Another observation
Moreover, from Example 2, we observe that TextGAIL seems to generate more reasonable and logical endings than MLE.
% Explain
We suspect that TextGAIL's discriminator has used some latent information to distinguish between the real and generated samples.
In Section 6, we specifically analyze what the discriminator has learned to provide useful rewards to the generator.

\subsection{Ablation Studies}

\begin{table}[h]
    \centering
    \resizebox{0.47\textwidth}{!}{
        \begin{tabular}{lccc}
            \toprule
                                & Perplexity $\downarrow$ & BLEU-2 $\uparrow$ & Distinct-2 $\uparrow$ \\
            \midrule
            SeqGAN + Pre-train*        & 140.42                  & n/a            & n/a \\    
            \midrule
            TextGAIL            & 14.85                   & 16.23            & 9.50
            \\
            \quad w/o PPO* & 111.08 & n/a & n/a \\
            \quad w/o mix human demo & 132.63 & n/a & n/a \\
            \quad w/o Contrastive Discriminator  & 17.26 & 13.72 & 8.40 \\
            \quad w/o D pre-train & 16.94  & 15.14  & 9.14  \\
            \bottomrule
        \end{tabular}
    }
    \caption{Ablation studies results on CommenGEN. "w/o D pre-train" means randomly initialized discriminator. * indicates that the model diverges during the training.}
    \label{tab:ablation}
\end{table}

Each component's contribution in TextGAIL's performance is shown on the CommonGEN task with an ablation study in Table~\ref{tab:ablation}.
We use beam search with beam size four as the inference method.

Some previous text GANs involve architecture changes such as LeakGAN \cite{DBLP:conf/aaai/GuoLCZYW18} and RelGAN \cite{DBLP:conf/iclr/NieNP19}, it is hard to directly apply them on the Transformer-based \cite{DBLP:conf/nips/VaswaniSPUJGKP17} models. 
Also, most of them are not designed for conditional generation tasks.
Therefore, we only test incorporating pre-trained language models for SeqGAN.
The model fails to converge.
It is probably due to the large number of parameters that needs update and the high variance in gradients.
% Observation 1
This suggests that the optimization techniques of PPO and mixed human demonstrations are crucial for stably training the text GANs.

We test replacing the contrastive discriminator with a normal discriminator that classifies a singe input with sigmoid.
The BLEU-2 and Distinct-2 scores decreases significantly.
We suspect that when the discriminator can only see one single input without comparing against the real example, it would be harmful for conditional generation tasks, as even if the generator outputs a sentence better than the ground truth, the generator cannot receive the accurate reward signal.

Moreover, we experiment the discriminator without any pre-training. 
The performance drops as expected.
This suggests the importance of pre-training for TextGAIL to improve over the MLE method.
In Section 6, we further explore what the discriminator has learned during adversarial learning.
% However, we find that it is quite limited to only make the jud 
% In Section 6, we further explore what the discriminator has learned during adversarial learning.

% Perhaps clarify what "D no pre-train". I assume it means initialized randomly rather than from RoBERTa weights, correct?

\section{What Has the Discriminator Learned}

\begin{table}[h]
    \centering
            \resizebox{0.45\textwidth}{!}{
                \begin{tabular}{lcc}
                    \toprule
                    Models                   & Supervised?  & Accuracy(\%)    \\
                    \midrule
                    RoBERTa w/ extra data    & $\checkmark$ & $92.8\pm0.28$   \\
                    \midrule
                    GPT-2 + MLE              & $\times$     & 69.6 $\pm$ 0.35 \\
                    \midrule
                    TextGAIL D               & $\times$     & 79.1 $\pm$ 0.76 \\
                    TextGAIL D w/o pre-train & $\times$     & 51.2 $\pm$ 0.85 \\
                    \bottomrule
                \end{tabular}
            }
    \caption{Story Cloze Test results. "D" means the discriminator. TextGAIL's learned discriminator can classify the story ending with the correct commonsense.}
    \label{tab:analysis}
\end{table}

% Why doing this
We analyze the reward signal of the learned discriminator in TextGAIL, which is supposed to distinguish the real samples from the generated samples.
% What we do
We apply the learned discriminator in TextGAIL on a story ending classification task, Story Cloze Test, to identify story endings with the correct commonsense given the story prompt \cite{DBLP:journals/corr/MostafazadehCHP16}. 
% Explanation of this task
%The task is to choose the right story ending from the wrong ending given the story prompt. 
This task uses a different dataset from ROCStories but is in the similar domain.
We report the Story Cloze Test results in Table~\ref{tab:analysis}. 

TextGAIL's discriminator achieves $79.1\%$ accuracy. 
This suggests the learned discriminator provides meaningful rewards to the generator in the training process.
% What we do
We compare our learned discriminator against a RoBERTa classifier fine-tuned on the Story Cloze Test's training data to explore if adding more supervision affects the performance.
We find that the fine-tuned RoBERTa classifier achieves the best accuracy (92.8\%). 
Clearly, direct supervision improves the performance, but our zero-shot TextGAIL discriminator is not too far from the supervised model's performance.

Inspired by \citet{DBLP:journals/corr/abs-1806-02847}, we also construct another baseline, the GPT-2 fine-tuned on ROCStories data with MLE (not Story Cloze Test), which also does not require extra supervision.
The model selects the ending with a higher joint language model probability.
It reaches 69.6\% accuracy, which is significantly worse than TextGAIL discriminator (79.1\%). 
This result, in another way, suggests TextGAIL has better reward guidance than the MLE language model.

We also compare TextGAIL's Discriminator against the model without pre-training (from ablation study) to see how much pre-training contributes to the performance. 
The accuracy drops from $79.1\%$ to 51.2 \% . 
This finding suggests that TextGAIL's discriminator is relying on the information obtained from pre-training to select the correct story ending.

% The results

% Explaination

\section{Conclusion}

% Background
% Despite the success on image generation, GANs have not yet achieved considerable progress on text generation.
% One of the difficulties is the instability of applying GANs on discrete data.
% Also, we observe that another reason may be the lack of sufficient language priors for the discriminator.
% Instead of designing task-specific techniques, we inherit previous work and extend to pretrained language models and using GAIL and efficient use of human demonstrations to stabilize the training.

% There are many criticisms about the existing research of GANs on text generation.
% One reason is that previous works lack fair comparisons of both the quality and diversity under different temperatures.
% Another is that current pre-trained language models such as GPT-2 are in the lead of MLE based models.
We propose a generative adversarial imitation learning framework for text generation - TextGAIL, which leverages the large pre-trained language models. We extend the exploration of adversarial training on text generation by incorporating large scale pre-trained models.
We use a contrastive discriminator and proximal policy optimization to improve the stability of the generator's training. We incorporate a large-scale pre-trained language model, GPT2 in our framework as the generator.
We also use a pre-trained RoBERTa model to initialize the discriminator to provide reliable rewards to the imitation learning framework.
Experiment shows that TextGAIL can generate not only more diverse but more accurate and reasonable outputs, and the discriminator can provide meaningful reward signals in various unconditional and conditional text generation tasks.

\bibliography{references}

\newpage
\appendix
\onecolumn

\section{Appendix}

\subsection{Perception Score}

% Why we need Perception Score
Traditional n-gram matching metrics such as BLEU and ROUGE are not sufficient to evaluate open-ended text generation tasks.
More and more studies rely on using human evaluators to determine the final quality of generated text.
However, fully using human evaluation instead of automatic evaluation is expensive and inconvenient in practice.
% The community calls for a more general automatic evaluation metric that correlates with human evaluation scores.

% Introduce
We introduce \textit{Perception score}, which is a simple automatic evaluation metric for text generation built on the top of fake text detector.
We report the accuracy of detecting fake text as the Perception score.
% More detail
The assumption of this approach is that the pretrained language models such as BERT have acquired plenty of prior knowledge during pretraining.
We can use them to detect not only the statistical difference, but also the semantic errors of the generated text.
If it has low accuracy in detecting generated text, it means that the real and generated text have very similar distribution.
Also, to reduce variance, we only train the discriminator with one epoch on a fixed dataset, where the generated samples are acquired from greedy sampling for deterministic outputs.

% The drawback and future direction
However, the drawback for this method is that it relies on the researchers themselves to train this fake text detector
Its result can be manipulated easily and misleading.
It is better to use it as a self-rating rather than a metric for comparison.
However, while Perception score is still a preliminary version of the metric and has limitations, we believe it merits further exploration to fill in the blank of text evaluation metrics.

We use a large pretrained language model RoBERTa-base as a classifier to distinguish between the real and generated samples.
We train the classifier for one epoch on the validation set, and use greedy sampling to sample outputs from the generator.
Then we deploy the classifier to the test set, and report the accuracy as the final score.
The resulting score approximately reflects the similarity of the real and generated distributions.

\begin{table}[h]
    \small
    \centering
    \resizebox{0.45\textwidth}{!}{
        \begin{tabular}{l|c}
            \toprule % <-- Bottomrule here
            Models                               & Perception$\downarrow$     \\
            \midrule
            GPT-2 + MLE                          & 0.94 $\pm$ 0.0067          \\
            \midrule
            TextGAIL                             & \textbf{0.77} $\pm$ 0.0095 \\
            \quad w/ random init discriminator** & 0.95 $\pm$ 0.0093          \\
            \midrule
            Human                                & 0.50*                      \\
            \bottomrule
        \end{tabular}
    }
    \caption{Automatic evaluation results.
        % Temperature is set to be 0.8 for all models. 
        *Perception Score for human written endings is theoretically $0.5$. **The discriminator is initialized with random weights instead of RoBERTa-base.}
    \label{tab:automatic_evaluation}
\end{table}

\subsection{Dataset Details}

We use COCO Image Captions to generate the text part only, which is the same setting as the previous works.
EMNLP2017 News is a dataset about news articles.
CommonGEN is a dataset in which giving three words, the model use the three words to generate a sentence.
ROCStories is a dataset about generate story endings, in which a story prompt is given, and the model needs to generate the corresponding ending.
The statistics of each dataset is shown in the following table.

\begin{table}[H]
    \begin{center}
        \begin{tabular}{lcccc}
            \toprule
                                     & COCO     & EMNLP2017 News & CommonGEN    & ROCStories   \\
            \midrule
            Condition generation     & $\times$ & $\times$       & $\checkmark$ & $\checkmark$ \\
            \midrule
            Vocabulary size          & 4,681    & 12,972         & 12,941       & 20,937       \\
            \midrule
            Average length           & 11.36    & 27.03          & 10.97        & 10.59        \\
            \midrule
            \# of Training Samples   & 10,000   & 278,586        & 39,069       & 98,161       \\
            \midrule
            \# of Validation Samples & 0        & 0              & 4,018        & 1,871        \\
            \midrule
            \# of Test Samples       & 10,000   & 10,000         & 6,042        & 1,871        \\
            \bottomrule
        \end{tabular}
    \end{center}
    \caption{Datasets Statistics.
    }
    \label{tab:datasets}
\end{table}

\subsubsection{Dataset Examples}

To better understand what each dataset look like, we show some examples of each dataset.
For COCO and EMNLP NEWS, they are unconditional generation tasks.
For CommonGEN and ROCStories, they are conditional generation tasks with source text and target text.

\begin{table}[H]
    \begin{center}
        \begin{tabular}{p{13cm}}
            \toprule
            Text                                                                \\
            \midrule
            a man is sitting on a bench next to a bicycle.                      \\
            \\
            a car that seems to be parked illegally behind a legally parked car \\
            \\
            a blue boat themed bathroom with a life preserver on the wall       \\
            \\
            the bike has a clock as a tire.                                     \\
            \\
            a bathroom with a toilet, sink, and shower.                         \\
            \\
            a small closed toilet in a cramped space.                           \\
            \\
            a beautiful dessert waiting to be shared by two people              \\
            \\
            a cute kitten is sitting in a dish on a table.                      \\
            \\
            a crowd of people are waiting to get on a red bus.                  \\
            \\
            an office with desk computer and chair and laptop.                  \\
            \\
            \bottomrule
        \end{tabular}
    \end{center}
    \caption{COCO Dataset Examples}
\end{table}

\begin{table}[H]
    \begin{center}
        \begin{tabular}{p{13cm}}
            \toprule
            Text                                                                                                                                                                    \\
            \midrule
            However one leading education expert has warned that the group's plans are unlikely to receive the significant investment they would require.                           \\
            \\
            We got to a bus station in the evening, but our connection didn't leave until the following morning.                                                                    \\
            \\
            But I still believe it's a big difference, top 10 for a week or for a year or for multiple years, and getting there is not easy.                                        \\
            \\
            I was paid far too little to pick up a dead off of the ground and put it back in the box.                                                                               \\
            \\
            His son, Lincoln O'Barry, said immigration authorities had turned down his father's request to visit Japan using a tourist visa.                                        \\
            \\
            The meeting comes as government figures show that UK air strikes have risen from a monthly average of 29 between May and October last year to more than 80 in December. \\
            \\
            He died in January but his body had to be frozen until September while she saved up for the funeral.                                                                    \\
            \\
            I can hold my referendum any time up until the end of 2017 and it is much more important to get this right than to rush it.                                             \\
            \\
            Costa had gone in hard on Oscar with a tackle and the midfielder responded with an extremely heavy one of his own that caused the pair to square up.                    \\
            \\
            I'm also excited for her to meet our families and friends, and to have a tiny combination of my husband and me.                                                         \\
            \bottomrule
        \end{tabular}
    \end{center}
    \caption{EMNLP2017 NEWS Dataset Examples}
\end{table}

\begin{table}[H]
    \begin{center}
        \begin{tabular}{p{5cm}p{8cm}}
            \toprule
            Source                 & Target                                            \\
            \midrule
            clock square tower     & clock tower on top of the historic market square  \\
            \\
            chocolate sale stall   & chocolates for sale at a small stall              \\
            \\
            bank lake summer       & trees on a grassy bank by a lake in summer        \\
            \\
            cheese pizza sauce     & A plate of pizza with cheese and sauce.           \\
            \\
            display giraffe museum & A tall giraffe on display at a museum.            \\
            \\
            cheese salad tomato    & woman pours cheese in a salad of tomatoes and ham \\
            \\
            giraffes zebra zoo     & Two giraffes and a zebra in an outdoor zoo        \\
            \bottomrule
        \end{tabular}
    \end{center}
    \caption{CommonGEN Dataset Examples}
\end{table}

\begin{table}[H]
    \begin{center}
        \begin{tabular}{p{8cm}p{5cm}}
            \toprule
            Source                                                                                                                                                                                                                 & Target                                                    \\
            \midrule
            Lisa makes celebration cakes for a living. She was asked to make a grand five tier wedding cake. She is very excited and a bit nervous about this. She outdoes herself and makes the most beautiful wedding cake ever. & Afterward, Lisa was swamped for cake orders for weddings. \\
            \\
            My husband loves tv. He watches it too much. Watching tv can cause bad eyesight. my husband had problems with his eyes.                                                                                                & He should stop watching so much TV.                       \\
            \\
            I tried going to the park the other day. The weather seemed nice enough for a walk. Within minutes of getting there I started sneezing. My eyes were watery and it was hard to breathe.                                & My allergies were too bad and I had to go back home.      \\
            \\
            Sara had lost her cat. She was so sad! She put up signs all over the neighborhood. Then a wonderful thing happened.                                                                                                    & Somebody found her cat.                                   \\
            \\
            Mary liked jumping rope. She would do it all the time. It kept her in good shape. She always had energy.                                                                                                               & Mary was glad she learned to jump rope.                   \\
            \bottomrule
        \end{tabular}
    \end{center}
    \caption{ROCStories Dataset Examples}
\end{table}

\subsection{Training Details}

All experiments run on a 2080Ti with 11GB GPU memory.
Most of the experiments can be completed in one day.
TextGAIL is about 4~5 times slower in training than the MLE-based model.
The primary reason is that Transformer's decoding is currently much slower than the training, which can be computed in parallel.
However, there is more and more research focusing on how to speed up the decoding of Transformer, which can alleviate the problem in the future.

\subsubsection{Hyper-parameters}

We show the hyper-parameters used for each task during the training.
\begin{table}[H]
    \begin{center}
        \begin{tabular}{lcccc}
            \toprule
                                & COCO & EMNLP & CommonGEN & ROCStories \\
            \midrule
            batch size          & 32   & 8     & 32        & 16         \\
            sample batch size   & 32   & 8     & 32        & 16         \\
            ppo buffer size     & 128  & 128   & 128       & 128        \\
            ppo mini batch size & 16   & 8     & 8         & 8          \\
            ppo epoch           & 1    & 1     & 1         & 1          \\
            ppo epsilon         & 0.2  & 0.2   & 0.2       & 0.2        \\
            mix human ratio     & 0.3  & 0.3   & 0.3       & 0.3        \\
            learning rate       & 1e-5 & 1e-5  & 1e-5      & 1e-5       \\
            warm-up steps       & 100  & 200   & 100       & 100        \\
            \bottomrule
        \end{tabular}
    \end{center}
    \caption{Hyper-parameters used for training.}
\end{table}
The sample batch size is determined by the GPU memory. The ratio between the PPO buffer size and ppo mini batch controls the degree of update in a single step.

\subsection{Generated Examples}

We have listed some of the TextGAIL's generations here. For more samples, we have provided the generated outputs used for evaluation in the supplementary material.

\begin{table}[H]
    \begin{center}
        \begin{tabular}{p{13cm}}
            \toprule
            Text                                                              \\
            \midrule
            a small plane is flying through the air.                          \\
            \\
            a crowd of people are walking in the street                       \\
            \\
            a giraffe crossing a lake in the wild                             \\
            \\
            a group of people standing next to bicycles.                      \\
            \\
            a small dog in a white shirt standing in front of a shop.         \\
            \\
            a group of people are sitting on a bench.                         \\
            \\
            a man riding a motorcycle down the street with a silver helmet    \\
            \\
            a couple of men sitting on a bench looking at a tv.               \\
            \\
            a black and white photograph of an airplane parked on the runway. \\
            \\
            a black and white photo of a person with a motorcycle             \\
            \\
            \bottomrule
        \end{tabular}
    \end{center}
    \caption{COCO Generated Samples}
\end{table}

\begin{table}[H]
    \begin{center}
        \begin{tabular}{p{13cm}}
            \toprule
            Text                                                                                                                                                             \\
            \midrule
            The US economy grew at a 4. 1 per cent annual rate in the third quarter, the fastest pace since the third quarter of 2011, according to official data.           \\
            \\
            For example, the time frame for a person's retirement is so long that they don't even realize it's coming.                                                       \\
            \\
            In general, women, men and children are less likely to be involved in the care of their doctors and hospitals, or in the building of new hospitals.              \\
            \\
            What we don't know is how much is actually being paid for the treatment and what percentage of that is actually being paid for the NHS's part.                   \\
            \\
            The report states that about 1, 000 police officers are needed to provide support for those with mental health issues.                                           \\
            \\
            It is a joint effort between the various government departments and the EU's ambassador to the UK.                                                               \\
            \\
            It will be a much bigger, much better experience for both of us to learn that about ourselves and our experience in life as well as what we do for our families. \\
            \\
            If there are no suspects, an investigation will be carried out and the government will release the information at the end of March.                              \\
            \\
            I've been trying to do my best for the past couple of months but I'm still not happy with the situation.                                                         \\
            \\
            "More than two million people have been displaced from the city and about 6, 500 have fled the fighting in Aleppo, according to the UN                           \\
            \bottomrule
        \end{tabular}
    \end{center}
    \caption{EMNLP2017 NEWS Generated Samples}
\end{table}

\begin{table}[H]
    \begin{center}
        \begin{tabular}{p{5cm}p{8cm}}
            \toprule
            Source                   & Target                                                                  \\
            \midrule
            field look stand         & field looks like a soccer field with a few soccer players standing      \\
            \\
            dance kid room           & Someone dances down a hallway and then slides down a room full of kids. \\
            \\
            hair look sit            & A man looks down on a dog sitting on a long chair.                      \\
            \\
            climb equipment mountain & mountain with equipment climbing a mountain                             \\
            \\
            costume dancer stage     & a stage dressed dancer in a gown costume                                \\
            \\
            arm sit stretch          & a woman is stretched out by the arms of a man in yoga pose.             \\
            \\
            ball bounce time         & time on the ball bounces back and forth on a white background           \\
            \bottomrule
        \end{tabular}
    \end{center}
    \caption{CommonGEN Generated Samples}
\end{table}

\begin{table}[H]
    \begin{center}
        \begin{tabular}{p{8cm}p{5cm}}
            \toprule
            Source                                                                                                                                                                                                                                   & Target                                                                     \\
            \midrule
            Paul loves playing video games. His computer was having trouble keeping up with the latest ones. He decided to upgrade. After a while his entire computer was practically new.                                                           & Now Paul can play all the latest games with his friends.                   \\
            \\
            Cho wanted to do well in school. So he started a study group with his friends. He studied hard every weekend. His grades began to climb up.                                                                                              & He got an A on his study group.                                            \\
            \\
            Kate and Greg went to a little candy shop together. They looked around at their options and made their choice. They went up to the cashier and said what they wanted. The cashier, with unwashed hands, bagged the candy without gloves. & Kate and Greg had a hard time deciding on a chocolate chip cookie instead. \\
            \\
            Phil wanted to make steaks for dinner. He went to the local butcher shop to get some meat. The guy working behind the counter was really informative. He helped him pick out the perfect cuts.                                           & Phil ate the steak and was very satisfied.                                 \\
            \\
            Frank opened his present. It was a shiny red wagon. He started to cry. His father explained that he'd once gotten one as a gift.                                                                                                         & Frank was so happy when his father said it was his.                        \\
            \bottomrule
        \end{tabular}
    \end{center}
    \caption{ROCStories Generated Samples}
\end{table}

\end{document}

\maketitle

\appendix

\section{Dataset Details}

We use COCO Image Captions to generate the text part only, which is the same setting as the previous works.
EMNLP2017 News is a dataset about news articles.
CommonGEN is a dataset in which giving three words, the model use the three words to generate a sentence.
ROCStories is a dataset about generate story endings, in which a story prompt is given, and the model needs to generate the corresponding ending.
The statistics of each dataset is shown in the following table.

\begin{table}[H]
    \begin{center}
            \begin{tabular}{lcccc}
                \toprule
                                      & COCO     & EMNLP2017 News & CommonGEN    & ROCStories   \\
                \midrule
                Condition generation   & $\times$ & $\times$      & $\checkmark$ & $\checkmark$ \\
                \midrule
                Vocabulary size        & 4,681    & 12,972        & 12,941       & 20,937       \\
                \midrule
                Average length         & 11.36    & 27.03         & 10.97        & 10.59        \\
                \midrule
                \# of Training Samples & 10,000   & 278,586       & 39,069       & 98,161       \\
                \midrule
                \# of Validation Samples & 0 & 0 & 4,018 &  1,871 \\
                \midrule
                \# of Test Samples & 10,000 & 10,000 & 6,042 & 1,871 \\
                \bottomrule
            \end{tabular}
    \end{center}
    \caption{Datasets Statistics.
    }
    \label{tab:datasets}
\end{table}

\subsection{Dataset Examples}

To better understand what each dataset look like, we show some examples of each dataset.
For COCO and EMNLP NEWS, they are unconditional generation tasks.
For CommonGEN and ROCStories, they are conditional generation tasks with source text and target text.

\begin{table}[H]
    \begin{center}
            \begin{tabular}{p{13cm}}
                \toprule
                Text \\
                \midrule
                a man is sitting on a bench next to a bicycle. \\
                \\
                a car that seems to be parked illegally behind a legally parked car \\
                \\
                a blue boat themed bathroom with a life preserver on the wall \\
                \\
                the bike has a clock as a tire. \\
                \\
                a bathroom with a toilet, sink, and shower. \\
                \\
                a small closed toilet in a cramped space. \\
                \\
                a beautiful dessert waiting to be shared by two people \\
                \\
                a cute kitten is sitting in a dish on a table. \\
                \\
                a crowd of people are waiting to get on a red bus. \\
                \\
                an office with desk computer and chair and laptop. \\
                \\
                \bottomrule
            \end{tabular}
    \end{center}
    \caption{COCO Dataset Examples}
\end{table}

\begin{table}[H]
    \begin{center}
            \begin{tabular}{p{13cm}}
                \toprule
                Text \\
                \midrule
                However one leading education expert has warned that the group's plans are unlikely to receive the significant investment they would require. \\
                \\
                We got to a bus station in the evening, but our connection didn't leave until the following morning. \\
                \\
                But I still believe it's a big difference, top 10 for a week or for a year or for multiple years, and getting there is not easy. \\
                \\
                I was paid far too little to pick up a dead off of the ground and put it back in the box. \\
                \\
                His son, Lincoln O'Barry, said immigration authorities had turned down his father's request to visit Japan using a tourist visa. \\
                \\
                The meeting comes as government figures show that UK air strikes have risen from a monthly average of 29 between May and October last year to more than 80 in December. \\
                \\
                He died in January but his body had to be frozen until September while she saved up for the funeral. \\
                \\
                I can hold my referendum any time up until the end of 2017 and it is much more important to get this right than to rush it. \\
                \\
                Costa had gone in hard on Oscar with a tackle and the midfielder responded with an extremely heavy one of his own that caused the pair to square up. \\
                \\
                I'm also excited for her to meet our families and friends, and to have a tiny combination of my husband and me. \\
                \bottomrule
            \end{tabular}
    \end{center}
    \caption{EMNLP2017 NEWS Dataset Examples}
\end{table}

\begin{table}[H]
    \begin{center}
            \begin{tabular}{p{5cm}p{8cm}}
                \toprule
                Source & Target \\
                \midrule
                clock square tower & clock tower on top of the historic market square \\
                \\
                chocolate sale stall & chocolates for sale at a small stall \\
                \\
                bank lake summer & trees on a grassy bank by a lake in summer \\
                \\
                cheese pizza sauce & A plate of pizza with cheese and sauce. \\
                \\
                display giraffe museum & A tall giraffe on display at a museum. \\
                \\
                cheese salad tomato & woman pours cheese in a salad of tomatoes and ham \\
                \\
                giraffes zebra zoo & Two giraffes and a zebra in an outdoor zoo \\
                \bottomrule
            \end{tabular}
    \end{center}
    \caption{CommonGEN Dataset Examples}
\end{table}

\begin{table}[H]
    \begin{center}
            \begin{tabular}{p{8cm}p{5cm}}
                \toprule
                Source & Target \\
                \midrule
                Lisa makes celebration cakes for a living. She was asked to make a grand five tier wedding cake. She is very excited and a bit nervous about this. She outdoes herself and makes the most beautiful wedding cake ever. & Afterward, Lisa was swamped for cake orders for weddings. \\
                \\
                My husband loves tv. He watches it too much. Watching tv can cause bad eyesight. my husband had problems with his eyes. & He should stop watching so much TV. \\
                \\
                I tried going to the park the other day. The weather seemed nice enough for a walk. Within minutes of getting there I started sneezing. My eyes were watery and it was hard to breathe. & My allergies were too bad and I had to go back home. \\
                \\
                Sara had lost her cat. She was so sad! She put up signs all over the neighborhood. Then a wonderful thing happened. & Somebody found her cat. \\
                \\
                Mary liked jumping rope. She would do it all the time. It kept her in good shape. She always had energy. & Mary was glad she learned to jump rope. \\
                \bottomrule
            \end{tabular}
    \end{center}
    \caption{ROCStories Dataset Examples}
\end{table}

\section{Training Details}

All experiments run on a 2080Ti with 11GB GPU memory.
Most of the experiments can be completed in one day.
TextGAIL is about 4~5 times slower in training than the MLE-based model.
The primary reason is that Transformer's decoding is currently much slower than the training, which can be computed in parallel.
However, there is more and more research focusing on how to speed up the decoding of Transformer, which can alleviate the problem in the future.

\subsection{Hyper-parameters}

We show the hyper-parameters used for each task during the training.
\begin{table}[H]
    \begin{center}
            \begin{tabular}{lcccc}
                \toprule
                & COCO & EMNLP & CommonGEN & ROCStories \\
                \midrule
                batch size & 32 & 8 & 32 & 16 \\
                sample batch size & 32 & 8 & 32 & 16 \\ 
                ppo buffer size & 128 & 128 & 128 & 128 \\
                ppo mini batch size & 16 & 8 & 8 & 8 \\
                ppo epoch & 1 & 1 & 1 & 1 \\
                ppo epsilon & 0.2 & 0.2 & 0.2 & 0.2 \\
                mix human ratio & 0.3 & 0.3 & 0.3 & 0.3 \\
                learning rate & 1e-5 & 1e-5 & 1e-5 & 1e-5 \\
                warm-up steps & 100 & 200 & 100 & 100 \\
                \bottomrule
            \end{tabular}
    \end{center}
    \caption{Hyper-parameters used for training.}
\end{table}
The sample batch size is determined by the GPU memory. The ratio between the PPO buffer size and ppo mini batch controls the degree of update in a single step.

\section{Generated Examples}

We have listed some of the TextGAIL's generations here. For more samples, we have provided the generated outputs used for evaluation in the supplementary material.

\begin{table}[H]
    \begin{center}
            \begin{tabular}{p{13cm}}
                \toprule
                Text \\
                \midrule
                a small plane is flying through the air. \\
                \\
                a crowd of people are walking in the street \\
                \\
                a giraffe crossing a lake in the wild \\
                \\
                a group of people standing next to bicycles. \\
                \\
                a small dog in a white shirt standing in front of a shop. \\
                \\
                a group of people are sitting on a bench. \\
                \\
                a man riding a motorcycle down the street with a silver helmet \\
                \\
                a couple of men sitting on a bench looking at a tv. \\
                \\
                a black and white photograph of an airplane parked on the runway. \\
                \\
                a black and white photo of a person with a motorcycle \\
                \\
                \bottomrule
            \end{tabular}
    \end{center}
    \caption{COCO Generated Samples}
\end{table}

\begin{table}[H]
    \begin{center}
            \begin{tabular}{p{13cm}}
                \toprule
                Text \\
                \midrule
                The US economy grew at a 4. 1 per cent annual rate in the third quarter, the fastest pace since the third quarter of 2011, according to official data.\\
                \\
                For example, the time frame for a person's retirement is so long that they don't even realize it's coming. \\
                \\
                In general, women, men and children are less likely to be involved in the care of their doctors and hospitals, or in the building of new hospitals. \\
                \\
                What we don't know is how much is actually being paid for the treatment and what percentage of that is actually being paid for the NHS's part. \\
                \\
                The report states that about 1, 000 police officers are needed to provide support for those with mental health issues. \\
                \\
                It is a joint effort between the various government departments and the EU's ambassador to the UK. \\
                \\
                It will be a much bigger, much better experience for both of us to learn that about ourselves and our experience in life as well as what we do for our families. \\
                \\
                If there are no suspects, an investigation will be carried out and the government will release the information at the end of March. \\
                \\
                I've been trying to do my best for the past couple of months but I'm still not happy with the situation. \\
                \\
                "More than two million people have been displaced from the city and about 6, 500 have fled the fighting in Aleppo, according to the UN \\
                \bottomrule
            \end{tabular}
    \end{center}
    \caption{EMNLP2017 NEWS Generated Samples}
\end{table}

\begin{table}[H]
    \begin{center}
            \begin{tabular}{p{5cm}p{8cm}}
                \toprule
                Source & Target \\
                \midrule
               field look stand & field looks like a soccer field with a few soccer players standing \\
                \\
                dance kid room & Someone dances down a hallway and then slides down a room full of kids. \\
                \\
                hair look sit & A man looks down on a dog sitting on a long chair. \\
                \\
                climb equipment mountain & mountain with equipment climbing a mountain \\
                \\
                costume dancer stage & a stage dressed dancer in a gown costume \\
                \\
                arm sit stretch & a woman is stretched out by the arms of a man in yoga pose. \\
                \\
                ball bounce time & time on the ball bounces back and forth on a white background \\
                \bottomrule
            \end{tabular}
    \end{center}
    \caption{CommonGEN Generated Samples}
\end{table}

\begin{table}[H]
    \begin{center}
            \begin{tabular}{p{8cm}p{5cm}}
                \toprule
                Source & Target \\
                \midrule
                Paul loves playing video games. His computer was having trouble keeping up with the latest ones. He decided to upgrade. After a while his entire computer was practically new. & Now Paul can play all the latest games with his friends. \\
                \\
                Cho wanted to do well in school. So he started a study group with his friends. He studied hard every weekend. His grades began to climb up. & He got an A on his study group. \\
                \\
                Kate and Greg went to a little candy shop together. They looked around at their options and made their choice. They went up to the cashier and said what they wanted. The cashier, with unwashed hands, bagged the candy without gloves. & Kate and Greg had a hard time deciding on a chocolate chip cookie instead. \\
                \\
                Phil wanted to make steaks for dinner. He went to the local butcher shop to get some meat. The guy working behind the counter was really informative. He helped him pick out the perfect cuts. & Phil ate the steak and was very satisfied. \\
                \\
                Frank opened his present. It was a shiny red wagon. He started to cry. His father explained that he'd once gotten one as a gift. & Frank was so happy when his father said it was his. \\
                \bottomrule
            \end{tabular}
    \end{center}
    \caption{ROCStories Generated Samples}
\end{table}